\documentclass[letterpaper]{article} 
\usepackage{aaai24}  
\usepackage{times}  
\usepackage{helvet}  
\usepackage{courier}  
\usepackage[hyphens]{url}  
\usepackage{graphicx} 
\usepackage{natbib}  
\usepackage{caption} 
\usepackage{algorithm}
\usepackage{algorithmic}
\usepackage{multirow}
\usepackage{subfigure}
\usepackage{booktabs}
\usepackage{newfloat}
\usepackage{listings}
\DeclareCaptionStyle{ruled}{labelfont=normalfont,labelsep=colon,strut=off} 
\floatstyle{ruled}
\newfloat{listing}{tb}{lst}{}
\floatname{listing}{Listing}
\title{How to Evaluate the Generalization of Detection? A Benchmark for Comprehensive Open-Vocabulary Detection}
\author{
    Yiyang Yao\textsuperscript{\rm 1}, 
    Peng Liu\textsuperscript{\rm 2}, 
    Tiancheng Zhao\textsuperscript{\rm 3}, 
    Qianqian Zhang\textsuperscript{\rm 2}, 
    Jiajia Liao\textsuperscript{\rm 3}, 
    Chunxin Fang\textsuperscript{\rm 3}, 
    Kyusong Lee\textsuperscript{\rm 3}, 
    Qing Wang\textsuperscript{\rm 1}
}
\affiliations{
    \textsuperscript{\rm 1} Northwestern Polytechnical University\\
    \textsuperscript{\rm 2} Linker Technology Research Co. Ltd\\
    \textsuperscript{\rm 3} Binjiang Institute of Zhejiang University \\
    yaoyiyang@mail.nwpu.edu.cn, \{liu\_peng,zhang\_qianqian\}@hzlh.com, \{tianchez,liaojiajia,fangchunxin,kyusongl\}@zju-bj.com, qwang@nwpu.edu.cn
}

\usepackage{bibentry}

\begin{document}

\maketitle

\begin{abstract}
Object detection (OD) in computer vision has made significant progress in recent years, transitioning from closed-set labels to open-vocabulary detection (OVD) based on large-scale vision-language pre-training (VLP). However, current evaluation methods and datasets are limited to testing generalization over object types and referral expressions, which do not provide a systematic, fine-grained, and accurate benchmark of OVD models' abilities. In this paper, we propose a new benchmark named OVDEval, which includes 9 sub-tasks and introduces evaluations on commonsense knowledge, attribute understanding, position understanding, object relation comprehension, and more. The dataset is meticulously created to provide hard negatives that challenge models' true understanding of visual and linguistic input. Additionally, we identify a problem with the popular Average Precision (AP) metric when benchmarking models on these fine-grained label datasets and propose a new metric called Non-Maximum Suppression Average Precision (NMS-AP) to address this issue. Extensive experimental results show that existing top OVD models all fail on the new tasks except for simple object types, demonstrating the value of the proposed dataset in pinpointing the weakness of current OVD models and guiding future research. Furthermore, the proposed NMS-AP metric is verified by experiments to provide a much more truthful evaluation of OVD models, whereas traditional AP metrics yield deceptive results. Data is available at \url{https://github.com/om-ai-lab/OVDEval}
\end{abstract}

\section{Introduction}
Open vocabulary detection (OVD) models have experienced rapid development in recent years, with numerous innovative techniques being introduced to the field. Novel models such as GLIP~\cite{li2022grounded}, Grounding DINO~\cite{liu2023grounding} and OmDet~\cite{zhao2022omdet} have introduced new vision-language learning methods such as modeling detection as visual grounding~\cite{kamath2021mdetr,li2022grounded}, pre-training with coarse image-text pairs~\cite{dou2022coarse}, and multi-task learning with a variety of detection tasks~\cite{zhao2022omdet}. 

As a result, for the first time, we can achieve strong zero-shot object detection (OD) on popular datasets such as COCO~\cite{lin2014microsoft}, even surpassing the performance of some of the supervised methods~\cite{liu2023grounding}. Users can simply use natural language to specify the desired targets and OVD models can detect the described targets on the fly, which opens doors for many new applications such as interactive image-editing~\cite{shen2023hugginggpt}, Augmented Reality~\cite{li2023otter} and robotics~\cite{shah2023lm}.

Meanwhile, current common approaches to evaluate OVD models include zero-shot/few-shot testing on OD dataset with common objects like COCO~\cite{lin2014microsoft}, OD dataset with long-tail objects like LVIS~\cite{gupta2019lvis}, grounding such as Flickr30K~\cite{plummer2015flickr30k} and referral expression comprehension (REC) such as RefCOCO~\cite{yu2016modeling}. These datasets were challenging for traditional OD research, but no longer serve as a challenging enough benchmark for future OVD methods for the following reasons:
\begin{itemize}
\item \textbf{Lack of systematic probing of model’s generalization ability:} An ideal OVD model should be able to understand the fine-grained semantics in the language prompt and align the language with visual features. Thus, it is required to probe the OVD model from various linguistic aspects such as object type, visual attributes, object relationship, etc., to quantify an OVD model's generalization to various degrees of prompt complexity.  
\item \textbf{Lack of hard negative for real-world usage:} Existing grounding and REC data assume the text prompt is paired with the image. The OVD model is only required to localize the entities mentioned in the caption without the need to discriminate against hard negatives. However, real-world usages command an OVD model to detect described object without knowing if the caption is related to the image at all.
\end{itemize}

To address the above issues, this paper introduces OVDEval to provide a comprehensive evaluation of OVD models and test their robustness against hard negatives. OVDEval is inspired by behavioral testing~\cite{ribeiro2020beyond,zhao2022vl}, and consists of 9 large datasets that cover 6 linguistic aspects: \textit{object}, \textit{proper noun}, \textit{attribute}, \textit{position}, \textit{relationship}, and \textit{negation}. All of the data annotations are carefully annotated by human experts to guarantee data quality. Additionally, these sub-datasets are meticulously crafted to ensure that all negative labels are hard. As a result, OVDEval is able to rigorously test a model's true understanding of a given aspect, preventing them from achieving high scores on a particular dimension by taking advantage of data bias. 

Besides the proposed dataset, this work also proposes a new evaluation metric named Non-Maximum Suppression Average Precision (NMS-AP). We identifies the \textit{The Inflated AP Problem} where even with high-quality hard negatives, a poor OVD model can still achieve a deceptive high AP score due to limitations on the calculation process of AP. The proposed NMS-AP is able to effectively resolve the Inflated AP Problem issue and offers a truthful evaluation of OVD models. 

We compared six strong baseline models on the proposed OVDEval dataset. Experimental results show that the current state-of-the-art (SOTA) OVD models only achieve strong results in simple object detection, and performance drop significantly on visual attribute understanding, commonsense knowledge and etc. This shows the significance to have a comprehensive and truthful benchmark to reveal the weakness of SOTA systems and guides the direction of future improvement. Analysis results also confirm the effectiveness of the proposed NMS-AP metric, whereas the conventional AP score is 30\% higher than the model's actual performance. Further analysis indicates that the current OVD model is only able to detect object types reliably and shows how OVD models can deceive conventional AP metrics by predicting multiple bounding boxes for each potential target object.
    
The contributions of our work are summarized as follows:
\begin{itemize}
    \item We introduce the first OVD evaluation benchmark that comprehensively tests model abilities across six linguistic aspects with complex language prompts and well-designed hard negatives.
    \item We identify the inflated AP problem that applies to any OVD model with traditional AP metric. 
    \item We propose NMS-AP, a novel evaluation metric that addresses the inflated AP score problem associated with traditional AP and we show NMS-AP provides a more accurate evaluation of OVD models' performance when dealing with fine-grained described detection.
    \item We show extensive experiment results that reveal the limitations of current SOTA OVD models and verify the effectiveness of the proposed metric.
\end{itemize}

\section{Related Work}
\textbf{Progression from Fixed Labels to Open Vocabulary Expressions:} Traditional object detectors, such as Faster R-CNN~\cite{ren2015faster} and YOLO~\cite{redmon2016you}, rely on a closed-set vocabulary and are trained on datasets like COCO~\cite{lin2014microsoft} and Pascal VOC~\cite{hoiem2009pascal} with predefined categories.

Over time, the number of labels increased, with Object365~\cite{shao2019objects365} introducing 365 labels and LVIS~\cite{gupta2019lvis} surpassing a thousand. Also, datasets like ODinW~\cite{li2022elevater} focus on wilderness objects with 35 different domains. V3Det~\cite{wang2023v3det} further broadened object detection capabilities across an extensive range of categories, paving the way for OVD. In addition to object detection, a growing body of research is dedicated to referral expression comprehension (REC) and visual grounding. REC focuses on identifying objects based on textual descriptions provided. Notable datasets in this area include RefCOCO~\cite{yu2016modeling}, PhraseCut~\cite{wu2020phrasecut}, Flickr30K~\cite{plummer2015flickr30k} and Visual Genome~\cite{krishna2017visual}. The Described Object Detection (DOD) introduced recently, combines the principles of object visual detection and REC, with the goal of detecting objects across various described categories. However, the above-mentioned datasets often lack hard negatives, which can lead to models detecting objects based on general terms rather than recognizing fine-grained details. Moreover, existing datasets have not investigated the model's ability to utilize common sense knowledge for detecting objects such as landmarks, logos, and celebrities.

\textbf{Endeavor for Systematic Model Evaluations:}
Benchmark scores often do not provide a comprehensive understanding of a model's capabilities, as they tend to present a superficial evaluation that can be difficult to interpret. Consequently, researchers have sought to scrutinize machine learning (ML) models with greater precision and granularity. In the realm of natural language processing (NLP), CheckList~\cite{ribeiro2020beyond} evaluates a wide range of linguistic competencies, revealing the limitations of numerous leading NLP models. For computer vision, the Vision CheckList~\cite{du2022vision} assists system developers in understanding a model's potential by introducing various transformation techniques to generate an extensive array of test samples. In the vision-language multimodal domain, VL-Checklist~\cite{zhao2022vl} serves as a framework for examining the proficiency of vision-language processing (VLP) models.

In the field of OD, studies often report conventional Average Precision (AP) scores. However, without an in-depth analysis, these scores can be challenging to understand. To address this limitation, we propose a novel evaluation approach that investigates a model's proficiency across clearly defined dimensions. Additionally, we introduce an evaluation metric designed to tackle the problem of deceptively high AP scores.

\section{OVDEval Benchmark}
The utilization of commonly employed OD datasets is associated with certain limitations. Firstly, evaluating OD performance solely based on AP across all labels in these datasets provides only a basic assessment. The specific capabilities of the model, such as accurately identifying object positions, have not been thoroughly evaluated. Moreover, in order to maintain linguistic label diversity and comprehensiveness, the distinctions between labels within the same dataset are typically coarse-grained and easily distinguishable. However, the OD task in the real world is much more challenging than merely detecting obvious objects or expressions. It is crucial to include hard negative samples that possess similar linguistic meanings but refer to different objects. Considering these concerns, we propose a new comprehensive benchmark dataset called OVDEval. 
OVDEval is divided into 9 sub-datasets, each focusing on evaluating the OD capabilities across 6 aspects: \textit{object}, \textit{proper noun}, \textit{attribute}, \textit{position}, \textit{relationship}, and \textit{negation}. 
The utilization of this benchmark dataset offers 3 significant benefits:

\begin{itemize}
    \item Detailed understanding of OD models: By evaluating OD models across different linguistic aspects, we can gain a more detailed understanding of their performance. This allows us to gain insights into the strengths and weaknesses of OD models, thereby facilitating the identification of areas for improvement.
    \item Commonsense understanding performance: OVDEval is specifically designed with linguistic queries, including commonsense knowledge-based labels, which enable us to assess the model's commonsense capabilities in the context of multimodal OVD. This evaluation sheds light on how well the model interprets knowledge.
    \item Fine-grained hard negative labels: we have carefully selected hard negative samples that conflict with the ground truth labels for each object, which provide a straightforward assessment of the model's performance in specific aspects.
\end{itemize}

\subsection{Dataset Description}

\begin{figure}[htb]
  \centering
  \includegraphics[width=8.6cm]{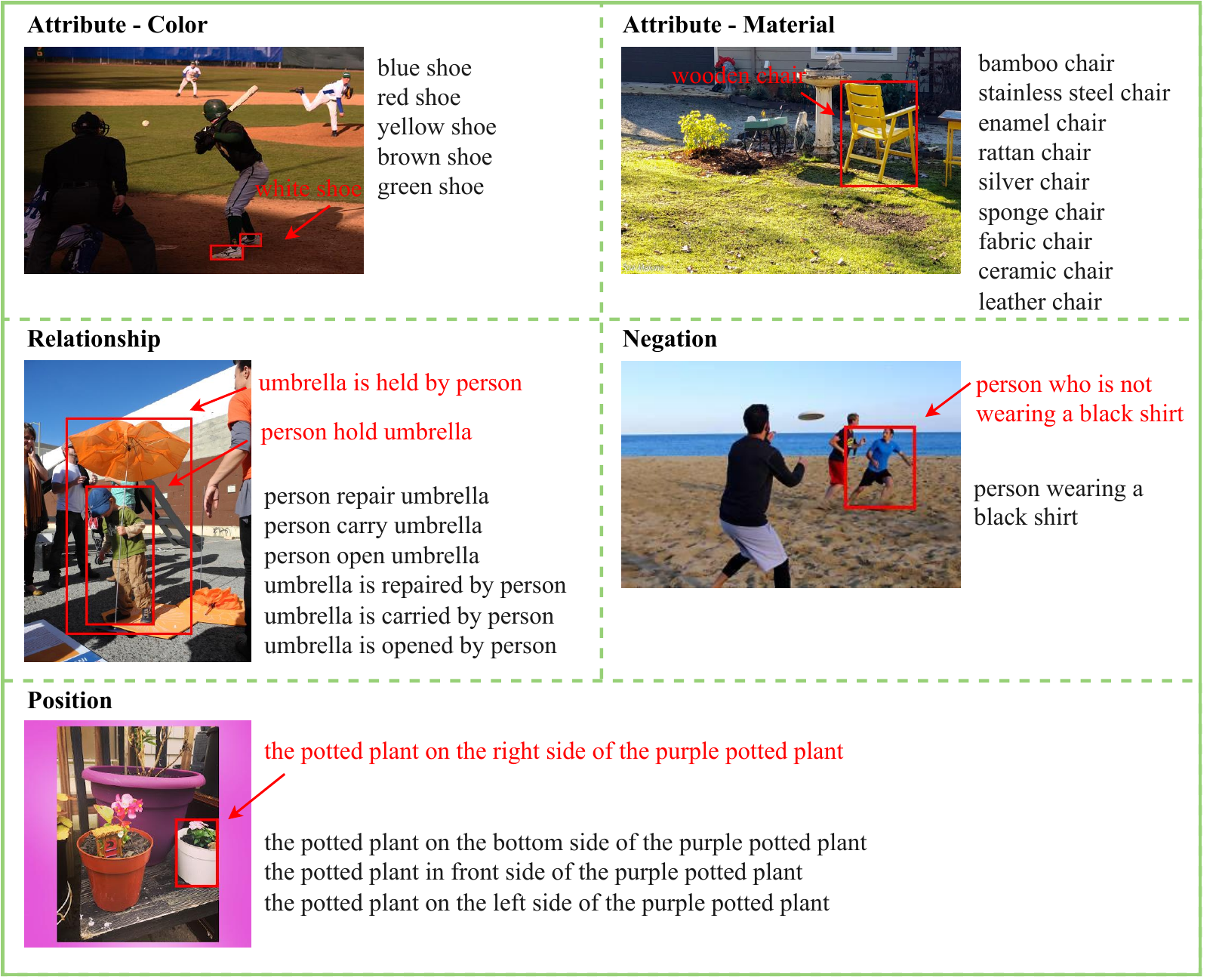}
  \caption{Detail-oriented dataset samples. Ground-truth labels are annotated with red color, and fine-grained hard negative samples are shown in black.}
  \label{fig_samples}
\end{figure}

We comprise the benchmark dataset from three viewpoints. Firstly, in line with existing datasets that primarily focus on evaluating the detection of common objects, we employ the COCO dataset to assess the models' general ability in this domain. Additionally, we aim to investigate the models' capacity to leverage external knowledge and common sense. Therefore, \textit{landmark}, \textit{logo}, and \textit{celebrity}, which require knowledge in both vision and language are implemented. Samples for the three aspects are shown in Figure \ref{knowledge}.  Furthermore, to delve into the models' proficiency in localizing fine-grained details, we divide the dataset into attributes (\textit{color} and \textit{material}), \textit{relationship}, \textit{position}, and \textit{negation} aspects.  Figure \ref{fig_samples} shows the detail-oriented dataset samples with corresponding  fine-grained hard negative samples, which essentially raise the detection difficulty. Finally, 9 sub-dataset across 6 aspects are collected and described as follows:
\begin{figure}[htb]
  \centering
  \includegraphics[width=8cm]{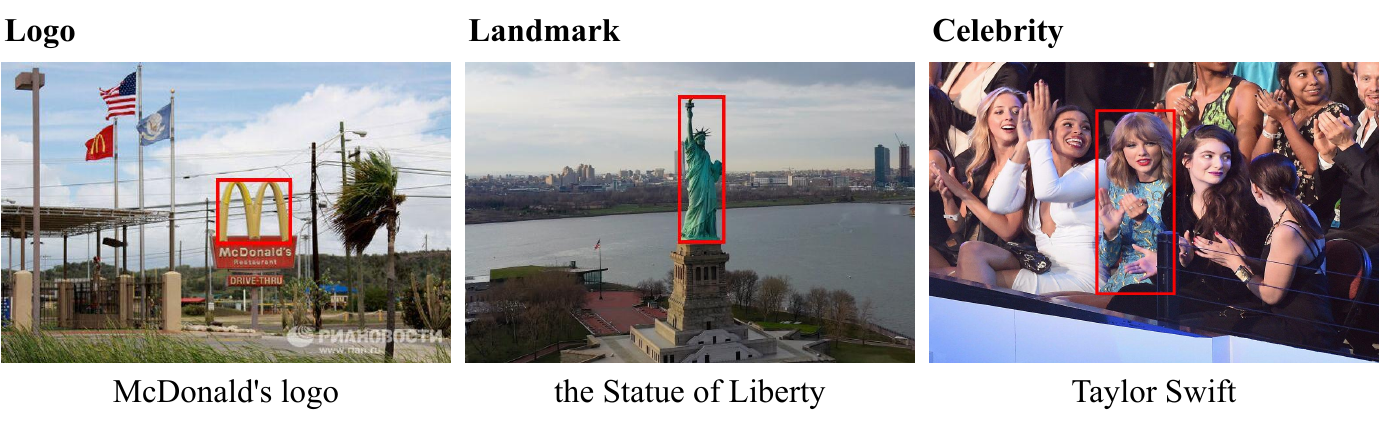}
  \caption{Samples of Proper noun datasets.}
  \label{knowledge}
\end{figure}
\begin{itemize}
    \item \textbf{\textit{Object}} is utilized to  evaluate the general capability in identifying common objects on the COCO Val 2017~\cite{lin2014microsoft}, which covers 80 common object categories.
    \item \textbf{\textit{Proper noun}} can unveil a model’s comprehension of commonsense knowledge including famous landmarks, renowned logos, and celebrities.
    \item \textbf{\textit{Attribute}} is used to assess OVD model's proficiency of distinguishing object characteristics. Specifically, \textit{color} and \textit{material} are employed as representing attribute aspects.
    \item  \textbf{\textit{Position}} aims to evaluate identifying specific objects among multiple visually similar items within a given image. The evaluation entails determining the target object based on the spatial relationships with other  described object expressions.
    \item \textbf{\textit{Relationship}} involves the examination of interactions between humans and other objects to comprehend both active and passive relationships among multiple objects.
    \item \textbf{\textit{Negation}} focuses on identifying objects expressed negatively, like spotting kitchen staff not wearing gloves. This checks the model's skill in detecting objects expressed in a negated context.
\end{itemize}

\subsection{Dataset Collection Process}

\subsubsection{Image Collection}
We collected varied images from three main sources. We used popular datasets, notably COCO and HICO~\cite{chao2018learning}. For evaluation, the COCO Val 2017 was directly used. For \textit{relationship}, the HICO dataset was the key source. After selecting the top-most frequent interaction label and excluding it, this selection process was repeated 10 times. We also changed active expressions to passive ones, ensuring two distinct labels for each sample image.

For the \textit{color} sub-dataset, we identified the top 50 objects from the visual genome (VG) dataset \cite{krishna2017visual}, labeling them using Oscar \cite{li2020oscar}. This enabled labeling objects from VG with colors. We concentrated on six object categories and six distinct colors, leading to 36 object-color combinations. Images were then randomly chosen from VG based on Oscar's labels.

For other datasets (\textit{landmark}, \textit{logo}, etc.), images came from the Laion-400m dataset \cite{schuhmann2021laion}. We began by identifying key terms for each subset. Using CLIP \cite{radford2021learning}, a top-tier image-text match model, images were sourced based on these keywords. To ensure variety, we crafted specific search prompts, considering context and diversity. For \textit{position} and \textit{negation}, we added terms like "multiple" to get images with several similar items.

\subsubsection{Hard Negative Labels}
We have implemented a novel approach that incorporates fine-grained hard negative labels for each linguistic aspect. These carefully selected hard negative labels are specifically designed to challenge the models and prevent them from achieving high scores on particular aspects without a genuine understanding. 

\begin{itemize}
\item For \textit{color}, variations in colors with the same object category are used to serve as negative labels. This approach exposes the OVD models to different color representations of the same object, thereby testing their ability to accurately distinguish and classify objects based on color. 
\item For \textit{material}, we maintain consistency in the object category while introducing variations in materials. 
\item For \textit{relationship}, we maintain the same subject and object entities but alter the verbs used to describe their relationship. 
\item In \textit{position}, we introduce changes in position words to serve as negative labels. For example, the words left can be replaced with right, above, under, front, back, and in. 
\item For \textit{negation}, we remove the word "not" from the positive labels as negative labels. 
\end{itemize}
The datasets with detail-oriented negatives essentially challenge the OVD models toward the advancement of object understanding in natural language processing.

\subsubsection{Manual Annotation}
To ensure the accuracy and reliability of the dataset, we engaged a team of OD annotation experts to manually annotate the collected images with a rigorous annotation process and a thorough quality inspection process. During the annotation process, any images with ambiguous labels were carefully identified and filtered out, guaranteeing the integrity of the final dataset. All the bounding boxes for the corresponding objects were annotated.

\begin{table*}[!htb]
\small
\begin{tabular}{@{}cccccccccc@{}}
\toprule
\multicolumn{1}{l}{\multirow{2}{*}{\textbf{}}} & \textbf{Object} & \multicolumn{2}{c}{\textbf{Attribute}} & \multicolumn{3}{c}{\textbf{Proper noun}}               & \multirow{2}{*}{\textbf{Relationship}} & \multirow{2}{*}{\textbf{Position}} & \multirow{2}{*}{\textbf{Negation}} \\ \cmidrule(lr){2-7}
\multicolumn{1}{l}{}                  & \textbf{COCO}          & \textbf{Color}   & \textbf{Material}   & \textbf{Landmark} & \textbf{Logo} & \textbf{Celebrity} &                                        &                                    &                                    \\ \midrule
\textbf{Images}                       & 5,000                    & 1,170            & 2,124               & 1,533             & 1,935         & 2,244              & 2,169                                  & 2,109                              & 1,858                              \\
\textbf{Bboxes}                       & 36,781                    & 3,421            & 5,358               & 1,709             & 2,329         & 2,244              & 8,190                                  & 2,150                              & 3,785                              \\
\textbf{Labels}                       & 80                   & 36               & 90                  & 9                 & 9             & 10                 & 319                                    & 7,301                              & 2,414                              \\
\textbf{Avg. negative labels}         & -                    & 5.01             & 8.73                & 8.00              & 8.00          & 9.00               & 7.65                                   & 3.06                               & 1.00                               \\
\textbf{Avg. label tokens}            & 6.03                     & 11.56            & 11.01               & 14.37             & 11.24         & 12.15              & 24.14                                  & 47.08                              & 27.34                              \\
\textbf{Avg. label words}             & 1.10                   & 2.00             & 2.03                & 2.66              & 2.00          & 2.12               & 4.48                                   & 9.67                               & 5.35                               \\ \bottomrule
\end{tabular}
\caption{Statistics of OVDEval for the 9 sub-datasets. OVDEval provides fine-grained annotations with hard negatives.}
\label{data_statistic}
\end{table*}

\subsection{Statistics}
As shown in Table~\ref{data_statistic}, the full OVDEval dataset comprises 9 distinct sub-datasets, collectively offering a total of 20K high-quality images accompanied by 3K meticulously annotated labels. The statics of each sub-dataset is provided in Table~\ref{data_statistic}. Notably, each sub-dataset encompasses a range of 1K to 5K images, ensuring the diversity and representatives of samples. While some sub-datasets feature proper nouns with a limited number of labels, it is important to highlight that all other sub-datasets can be considered as open set labels. Moreover, these sub-datasets incorporate extremely hard negative labels, further pushing the boundaries of model performance and evaluation. The inclusion of open set labels and hard negative labels within the majority of the sub-datasets enhances the dataset's realism and reflects the complexities encountered in real-world scenarios.

\section{The Proposed Evaluation Metric}
\subsection{The Inflated AP Problem}
AP is defined as the area under the precision-recall curve. This metric evalautes a model's performance by considering the trade-off between precision and recall. Recent OD research has predominantly used the COCO AP as the major benchmark metric. In the COCO mean Average Precision (mAP) calculation, a 101-point interpolated AP definition is utilized. Specifically, for COCO, AP is determined as the average across multiple Intersection-over-Union (IoU) thresholds that determine a positive match. AP@[.5:.95] represents the average AP for IoU values ranging from 0.5 to 0.95, with a step size of 0.05. 

Considering a scenario where an OVD model demonstrates good zero-shot performance in detecting objects but totally does not understand contextual descriptions, the model can deceive traditional AP metrics and obtain a high score by generating multiple predicted bounding boxes for the target object with all candidate labels. Assuming an image with 2 annotated ground-truth instances, which are labeled as \textit{red car} and \textit{blue car}, respectively. Then, the aforementioned model predicts 4 bounding boxes, generating 2 for each target object and assigning both candidate labels to each box. The IoUs between predictions and corresponding ground-truth instances are assumed to be greater than 0.95. As a result, the precision and recall for each category can be derived using the following equation:
\begin{equation}
\small
    Precision = \frac{TP}{TP+FP} = \frac{1}{1+1} = 0.50
\end{equation}
\begin{equation}
\small
Recall = \frac{TP}{GTnum} = \frac{1}{1} = 1.0
\end{equation}
Here, TP is the number of correctly predicted instances for a specific category, while FP is the number of instances that were incorrectly predicted as belonging to that category. GTnum represents the total number of ground-truth instances in the image. In the given scenario, where the IoU of predictions is assumed to be greater than 0.95, we can ignore the AP calculation process for IoU values ranging from 0.5 to 0.95. Therefore, we can calculate the average AP of each category as 0.50. Consequently, the mAP would also be 0.50. In this case, the model deceives traditional AP metrics to get an mAP score of 0.50, even though it only detects the target objects without comprehending their descriptions. In this case, the conventional COCO AP metric demonstrates a vulnerability that we refer as \textit{The Inflated AP Problem}. During the stage of matching predictions with ground truth to count TP and FP, it only considers predictions that have the same label as the ground truth. As a result, OVD models can obtain inflated AP scores by simply predicting multiple bounding boxes on a single object with all possible labels.

The inflated AP problem can lead to misleading evaluations of OVD models, as it fails to capture the accuracy of the descriptive labels assigned to the objects. Therefore, it is essential to develop alternative evaluation metrics that consider both object detection and the understanding of linguistic descriptions to provide a more robust assessment of OVD models.

\begin{figure}[!h]
    \centering
    \subfigure[before NMS]{\includegraphics[width=4.1cm, height=3.3cm]{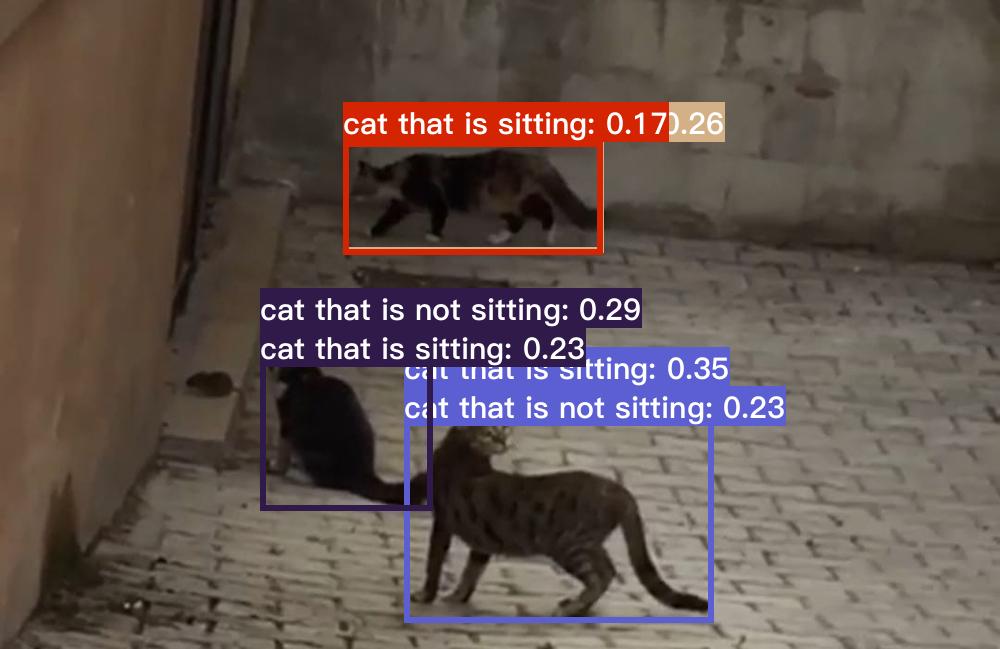}}
    \subfigure[after NMS]{\includegraphics[width=4.1cm, height=3.3cm]{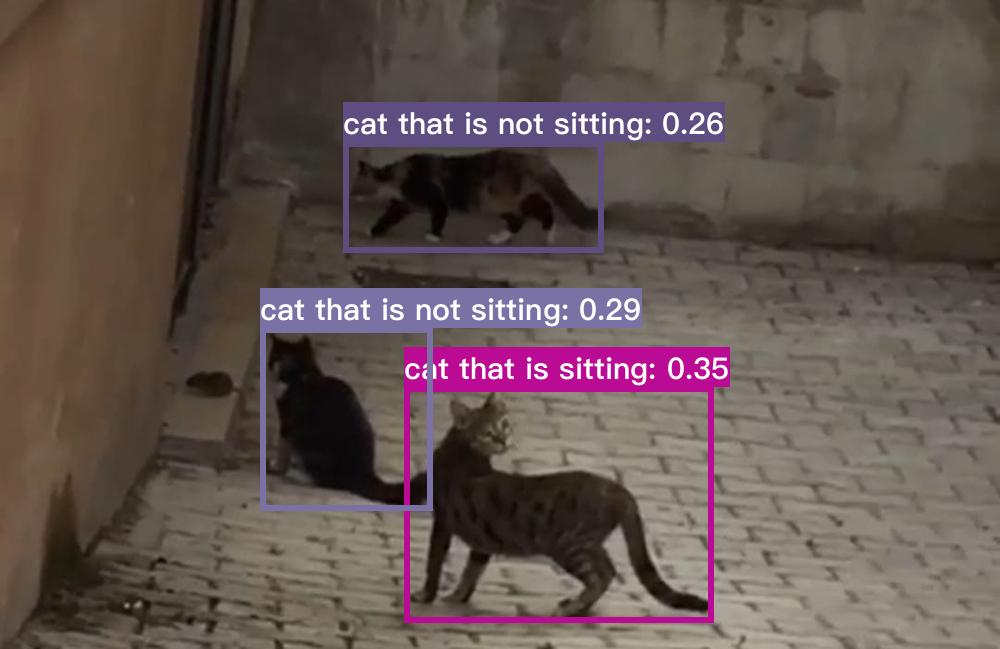}}

\caption{Examples of predictions from GLIP before and after class-ignored NMS, showing the limitation of current OVD models.}
\label{fig:examples}
\end{figure}

\subsection{A Simple Fix: NMS-AP}
To address the aforementioned issue, we propose a simple fix for the COCO AP metric, which we refer to as NMS-AP. It extends the traditional COCO AP metric by incorporating NMS~\cite{girshick2015fast}, a technique that is used in OD tasks to eliminate redundant bounding box predictions by selecting the most relevant ones based on their confidence scores and suppressing overlapping bounding boxes based on IoU. Specifics of NMS-AP are outlined below Algorithm \ref{alg:nms-ap}.

\begin{algorithm}
\small
  \caption{NMS-AP Metrics}
  \label{alg:nms-ap}
  \begin{algorithmic}[1]
    \renewcommand{\algorithmicrequire}{\textbf{Input:}}
    \REQUIRE $preds $: predictions
    \REQUIRE $GT $: ground-truth
    \STATE $ pickedPreds = keepPreds = []$
    \FOR{$k \ in\  GT $}
        \FOR{$p \ in\  preds $}
            \IF{IoU(\textit{p}, \textit{k}) \textgreater{0.5}}
                \STATE $pickedPreds = pickedPreds \cup p$
            \ELSE
                \STATE $keepPreds = keepPreds \cup p$
            \ENDIF
        \ENDFOR
        \STATE $keepPreds = keepPreds\cup \textit{C-NMS}(pickedPreds)$
    \ENDFOR
    
    \STATE $mAP = AP(keepPreds, GT)$
    \RETURN $mAP$
  \end{algorithmic}
\end{algorithm}

In NMS-AP, instead of considering only the prediction with the highest confidence score for each object, we apply a class-ignored NMS (C-NMS) to remove redundant predictions that match ground truth. To be specific, we employed class-ignored NMS on the predictions that exhibited an IoU \textgreater{0.5} when compared to the ground-truth instances. This ensures that multiple bounding boxes predicted for the same object are appropriately handled and only use the prediction with the highest confidence. In an ideal scenario with a flawless OVD model, it should predict bounding boxes with the correct label and the highest confidence score for each ground-truth instance. Consequently, the application of class-ignored NMS will solely remove false positives, ensuring that this model achieves a perfect score of 1.0. However, in the case of a subpar model that struggles to comprehend complex linguistic descriptions, the application of class-ignored NMS may lead to a decrease regarding true positives and NMS-AP scores (Figure~\ref{fig:examples}). This is because of the failure of accurately predict the bounding boxes that correspond to the ground-truth instances due to its limited understanding of the linguistic context. 

Note that NMS-AP is model-agnostic and can be apply to any OVD models. It simply takes a set of predictions and ground-truth bounding boxes, and removes overlapping predictions adjacent to the ground-truths. 

\section{Results and Analysis}
 
\begin{table*}[!ht]
\small
\centering
\begin{tabular}{@{}cccc@{}}
\toprule
\textbf{Model}         & \textbf{Pre-train Data}                          & \textbf{Backbone} & \textbf{Params} \\ \midrule
\textbf{Detic}         & ImageNet-21K,COCO,LVIS                           & Swin-B            & 141.6M          \\
\textbf{MDETR}         & VG,Flickr30k,COCO image-text pairs               & ResNet-101        & 185M            \\
\textbf{GLIP}          & FourODs,GoldG,CC3M+12M,SBU                       & Swin-L            & 430.42M         \\
\textbf{FIBER}         & Flickr30k, MixedNoCOCO, O365                     & Swin-B            & 252.06M         \\
\textbf{OmDet}         & O365,GoldG,PhraseCut,HOI-A,VAW,RefCOCO       & ConvNext-B        & 241.5M          \\ 
\textbf{Grounding DINO} & COCO,O365,GoldG,Cap4M,OpenImage,ODinW-35,RefCOCO & Swin-B            & 232.9M          \\ \bottomrule
\end{tabular}
\caption{ The relevant information of different models include pre-train data, backbone, and parameters.}
\label{tab:modelinfo}
\end{table*}

We conducted experiments on 9 datasets across 6 aspects using several leading publicly available models: Detic~\cite{zhou2022detecting}, MDETR~\cite{kamath2021mdetr}, GLIP~\cite{li2022grounded}, FIBER~\cite{dou2022coarse}, OmDet~\cite{zhao2022omdet} and Grounding DINO~\cite{liu2023grounding}. We provide these detailed model information such as pretraining data, backbone, and the number of parameters (Table~\ref{tab:modelinfo}).

\subsection{Main Results on NMS-AP on OVDEval}

\begin{table*}[!ht]
\small
\centering
\begin{tabular}{@{}cccccccccccccc@{}}
\toprule
\textbf{Aspects}    & \textbf{Sub-datasets} & \textbf{GLIP}      & \textbf{FIBER} & \textbf{Grounding DINO} & \textbf{Detic}       & \textbf{MDETR} & \textbf{OmDet}   \\ 
\cmidrule(l){3-8}  &      & NMS-AP/AP        & NMS-AP/AP      & NMS-AP/AP      & NMS-AP/AP      & NMS-AP/AP      & NMS-AP/AP   \\ \midrule
\textbf{Object}    & COCO      & 48.90 / 51.30    & 46.80 / 49.30        & 52.50$^*$ / 55.30$^*$      & 45.30$^*$ / 45.80$^*$   & 1.60 / 3.20          & \textbf{54.68} / 57.50         \\ 
\midrule & Logo     & 10.20 / 17.61         & 6.30 / 9.05      & \textbf{10.30} / 14.60    & 9.60 / 9.60  & 0.90 / 4.60      & 6.10 / 11.00         \\
 & Landmark     & 20.30 / 36.36    & 11.00 / 16.99     & 15.10 / 23.40    & \textbf{30.00} / 30.08 & 1.80 / 7.80     & 26.30 / 32.38      \\
 & Celebrity    & \textbf{4.60} / 8.24    & 0.80 / 3.31     & 0.70 / 2.00      & 0.00 / 0.00      & 1.10 / 4.80     & 1.80 / 6.36      \\
\textbf{Proper Noun}      & Avg     & 11.70 / 20.74    & 6.03 / 9.78   & 8.70 / 13.33    & \textbf{13.20} / 13.23  & 1.27 / 5.73     & 11.40 / 16.58           \\
\midrule  & Color    & 3.70 / 6.70     & 6.80 / 9.40    & 9.40 / 12.41     & 3.90 / 4.14     & 3.10 / 7.30      & \textbf{22.90} / 24.56 \\
& Material    & 7.40 / 15.87     & 12.40 / 17.72     & 9.00 / 15.50   & 9.20 / 9.75       & 2.50 / 10.70     &  \textbf{16.30} / 22.59         \\
\textbf{Attribute}    & Avg         & 5.55 / 11.28         & 9.60 / 13.56     & 9.20 / 13.96             & 6.55 / 6.94            & 2.80 / 9.00      & \textbf{19.60} / 23.58  \\ \midrule
\textbf{Position}       &     & 30.90 / 48.10         & 34.30 / 48.20    & \textbf{67.50} / 77.40  & 12.20 / 14.40           & 34.00 / 48.80    &   21.20 / 47.75     \\ \midrule
\textbf{Relationship}   &     & 10.00 / 33.20         & 14.50 / 31.40     & 10.70 / 35.30             & 6.10 / 7.20            & 8.20 / 29.40     & \textbf{41.98} / 51.98  \\ \midrule
\textbf{Negation}       &      & 29.30 / 51.80         & 28.70 / 57.20     & \textbf{52.50} / 67.30  & 27.90 / 29.70          & 28.30 / 41.10     & 35.10 / 55.86       \\ \midrule \midrule
\textbf{Total Average}    &     & 18.37 / 29.91        & 17.96 / 26.95   & 25.30 / 33.69     & 16.02 / 16.74         & 9.06 / 17.52       & \textbf{25.86} / 39.15        \\ \bottomrule
\end{tabular}
\caption{The NMS-AP and traditional AP evaluation results (\%), * represents supervised score, otherwise it's zero-shot. Total average is averaged over the 9 subtasks.}
\label{tab:nms-applied_ap}
\end{table*}

\begin{figure}[!htb]
  \centering
  \includegraphics[width=7cm]{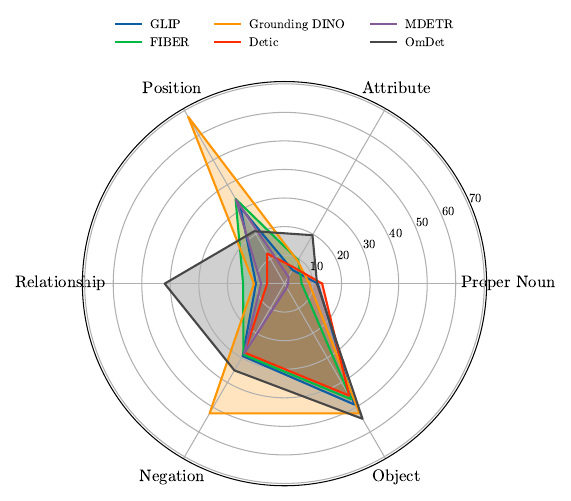}
  \caption{Radar chart of NMS-AP results on 6 aspects. Most models successfully worked on \textit{object} but failed on others.}
   \label{fig:radargraph}
\end{figure}

The experimental results, as presented in Table \ref{tab:nms-applied_ap}, show that current models generally perform satisfactorily on the \textit{object} task, with the exception of MDETR. This observation is consistent with earlier work that reported MDETR's low performance on the COCO dataset~\cite{cai2022x}. This indicates that most existing models possess strong capabilities in detecting objects. However, we observe that all current models exhibit poor performance on the \textit{logo}, \textit{landmark}, and \textit{celebrity} tasks in \textit{proper noun} aspect. Especially the NMS-AP values are close to 0\% in celebrity tasks. Notably, Detic demonstrates impressive results on the \textit{logo} and \textit{landmark} tasks, even without employing a complex fusion strategy, while its performance is relatively weak on tasks involving longer descriptions.

For datasets with hard negatives, the labels often involve some descriptions and require a more fine-grained linguistic understanding for models. We found that all models exhibit poor performance on \textit{color} and \textit{material} tasks. In contrast, OmDet performs more favorably overall on these tasks, largely due to its use of the VAW~\cite{Pham_2021_CVPR} dataset with attributes during pre-training. Meanwhile, the overall performance of existing models on the \textit{position}, \textit{relationship}, and \textit{negation} tasks is similar, with generally low NMS-AP values. This indicates that the current models have limited capability in handling tasks with fine-grained descriptions. However, we note that Grounding DINO significantly outperforms the other models in \textit{position} task. This can be attributed to its utilization of the RefCOCO dataset with orientation data during pre-training, which provides the model with specific knowledge related to the position and improves its performance on this task. Moreover, OmDet performs better than other models on the relationship task, which can be attributed to its use of the HOI-A~\cite{liao2020ppdm} dataset with relation attributes during pre-training, providing the model with specific knowledge related to the relationship and improving its performance on this task. While minor differences exist, all models display a similar trend when we represent the 6 aspects on a radar chart (Figure~\ref{fig:radargraph}). All models successfully worked on the common object task (\textit{object}). However, they all failed on the hard tasks from the proposed datasets, which require the use of external/commonsense knowledge and fine-grained localization ability. Therefore, it is evident that a dataset with fine-grained labels is necessary to establish a better benchmark to provide a clear optimization direction for improving the model's performance on challenging tasks.

\subsection{Comparing NMS-AP with Traditional AP}

To validate the Inflated AP problem, we performed the evaluation of traditional AP on our OVDEval dataset and compared it with the NMS-AP results.  Table~\ref{tab:nms-applied_ap} shows that the difference between NMS-AP and AP on classical OD datasets such as COCO is small, e.g., 52.50 vs. 55.30 for Grounding DINO because the probability of mutual exclusion of the predicted labels in this task is small and its impact on the AP calculation is negligible. On the other hand, the difference between NMS-AP and AP becomes much more significant for more difficult aspects including attribute, position and etc. For example, the \textit{relationship} AP of Grounding DINO decreased from 35.30 to 10.70. The above results confirmed that our hypothesis about the Inflated AP Problem exists for the compared OVD models. To visually illustrate our hypothesis and investigate the cause of the large NMS-AP and AP difference, we have plotted several bounding boxes obtained from the GLIP predictions, as illustrated in Figure \ref{fig:examples}.



From the examples depicted in Figure \ref{fig:examples}, it is evident that GLIP tends to generate multiple bounding boxes on the same object. Notably, the labels assigned to these bounding boxes are mutually exclusive. For instance, in the case of a cat, the predicted bounding boxes include both "cat that is sitting" and "cat that is not sitting". This inconsistency matches our hypothesis about the inflated score problem by deceiving traditional AP. That is although Grounding DINO has a poor performance in understanding \textit{negation} it can still obtain a high AP score. 

On the other hand, by employing our NMS-AP algorithm, we effectively retain only one bounding box with the highest confidence for each ground-truth instance while disregarding other false bounding boxes during the AP calculation. This approach helps mitigate the inflated AP problem caused by multiple bounding boxes. The decrease in scores that we observed earlier can be primarily attributed to models predicting the highest confidence on false labels, indicating a failure in comprehending fine-grained descriptions. Note that among all the models, Detic suffers the least from NMS-AP and AP difference because its model architecture already applies NMS internally to the region proposal network (RPN) that remove the duplicated boxes over the same region~\cite{zhou2022detecting}.

Therefore, utilizing NMS-AP to evaluate OVD models on our benchmark provides a more suitable approach for assessing their performance on intricate linguistic descriptions. This method helps address the limitations of the models and provides a more accurate evaluation metric.

\subsection{Limitations of Current OVD Models}
We have also noticed a recurring issue among all the OVD models, where they tend to generate multiple bounding boxes for the same object but assign inconsistent labels to them. Moreover, these predicted labels are often mutually exclusive, and it is worth mentioning that the predictions with the highest confidence scores are frequently incorrect. This issue is particularly pronounced in models with a large number of output bounding boxes, such as Grounding Dino. This observation further strengthens our previous hypothesis that the current models demonstrate exceptional performance in learning straightforward object tasks such as COCO. However, they encounter difficulties in comprehending the intricacies of detailed descriptions.

\begin{figure}[!h]
    \centering
    \subfigure[object aspect]{\includegraphics[width=3.9cm]{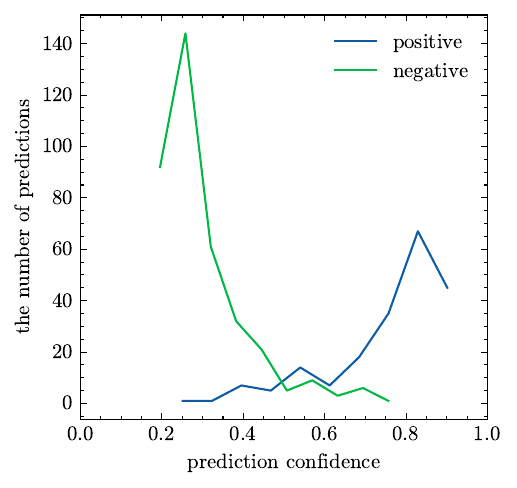}}
    \quad
    \subfigure[negation aspect]{\includegraphics[width=3.9cm]{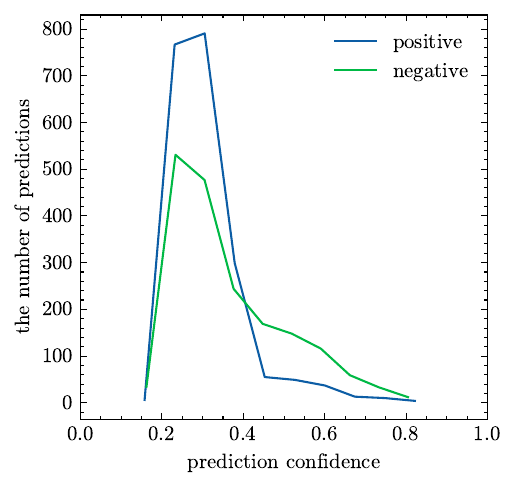}}

\caption{Distribution analysis of predicted confidence for object and negation aspects in GLIP}
\label{fig:coco-positive and negative}
\end{figure}

To further support our hypothesis, we plot the distribution of predicted confidence score for the \textit{object} and \textit{negation} aspects. Figure \ref{fig:coco-positive and negative} from GLIP illustrates the distribution of confidence scores. The distribution of \textit{object} is obtained from the model predictions on a subset of images in the COCO validation dataset. To calculate these distributions, we tally the number of positive and negative labels from the predictions that have an IoU greater than 0.9 with the ground truth.

Based on the results in Figure \ref{fig:coco-positive and negative}, it is clear that in the \textit{object} task, positive predictions tend to be spread out across the high confidence range, while negative predictions are mostly concentrated in the low confidence range. This indicates that most models have successfully learned to accurately identify objects. However, in the \textit{negation} task, the confidence distribution of positive and negative samples exhibits a similar trend. Meanwhile, the predictions predominantly appear in the low confidence region. These findings further support our hypothesis that existing models struggle to comprehend certain nuanced semantic information in fine-grained tasks.

\section{Conclusion}
This paper presents a novel benchmark OVDEval, testing the generalization of open-vocabulary detectors. We carefully create the dataset with challenging hard negatives and annotate 20K images with human experts. We also identified the Inflated AP problem for conventional AP calculation and introduce a new metric NMS-AP to deal with it. Our assessment validates the OVDEval's effectiveness in revealing the pros and cons of current SOTA open-vocabulary models. Lastly, OVDEval provides promising future research questions. How can we incorporate better training objectives so OVD models can acquire better discriminate abilities against hard negatives in both visual and linguistic input? What are the better pre-training data to inject more common sense knowledge in vision-language alignment? In summary, solving OVDEval is an important step for future general-purpose object detectors.

\section*{Acknowledgements}
This research is supported by National Key R\&D Program of China under grant (2022YFF0902600) and Key R\&D Program of Zhejiang under grant (2023C01048). Y.Y. Yao and Q. Wang are supported by NSFC under grant 62031023.

\bibliography{aaai24}
\section{Supplementary Material}

\subsection{Cases of predictions before and after applying class-ignored NMS}
\begin{figure*}[ht]
    \centering
    \subfigure[]{\includegraphics[width=6.3cm]{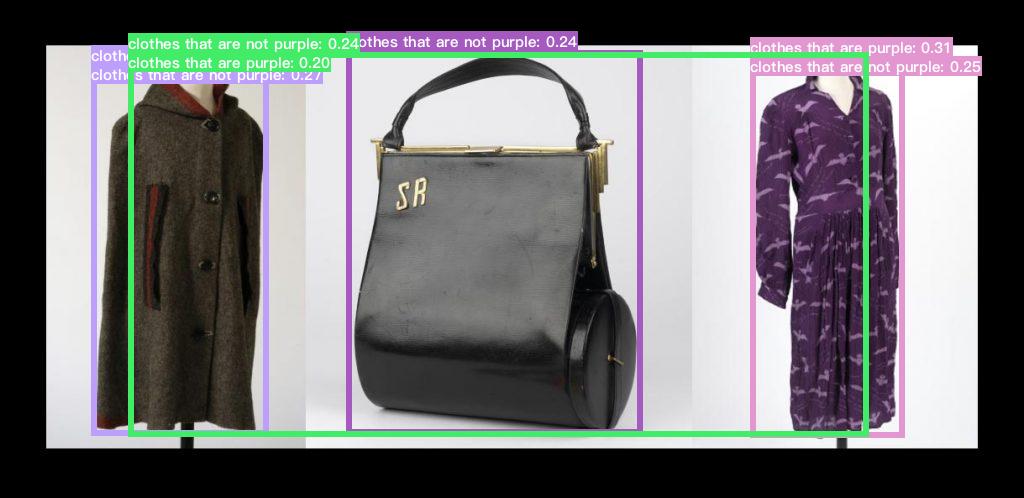}}
    \quad
    \subfigure[]{\includegraphics[width=6.3cm]{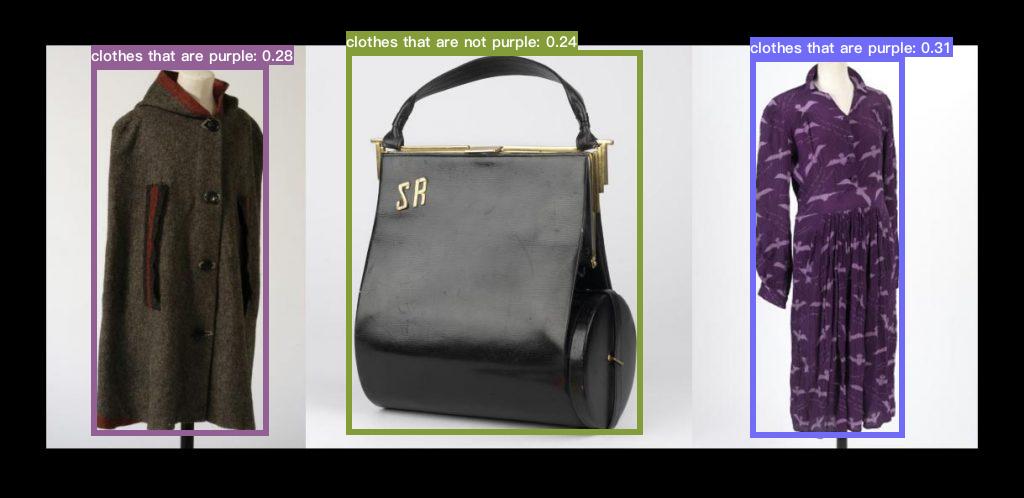}}
    \quad
    \quad
    \subfigure[]{\includegraphics[width=6.3cm]{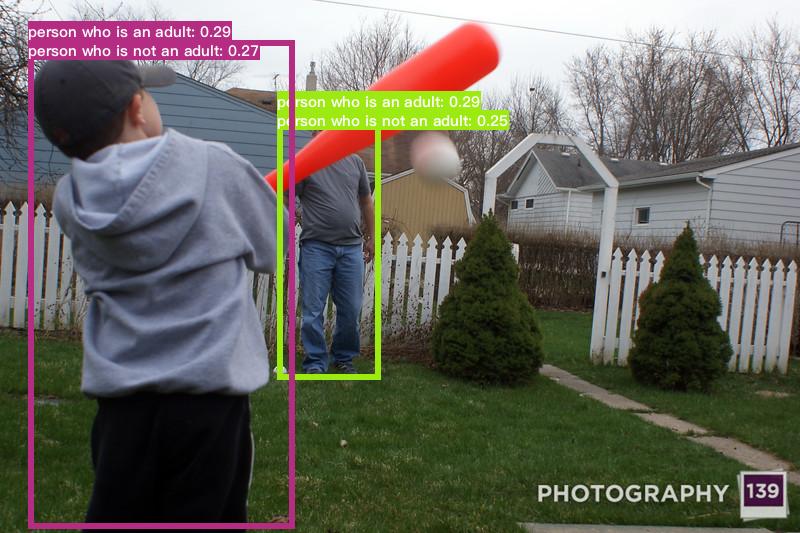}}
    \quad
     \subfigure[]{\includegraphics[width=6.3cm]{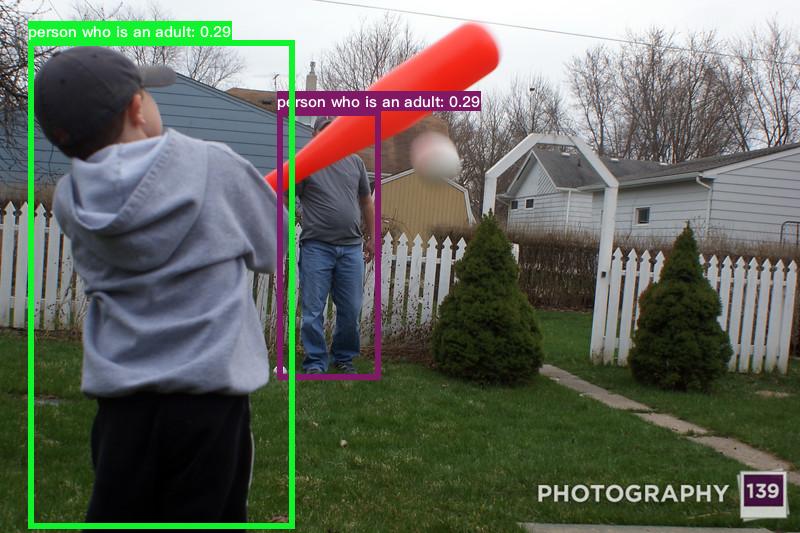}}
    \quad
    \subfigure[]{\includegraphics[width=6.3cm]{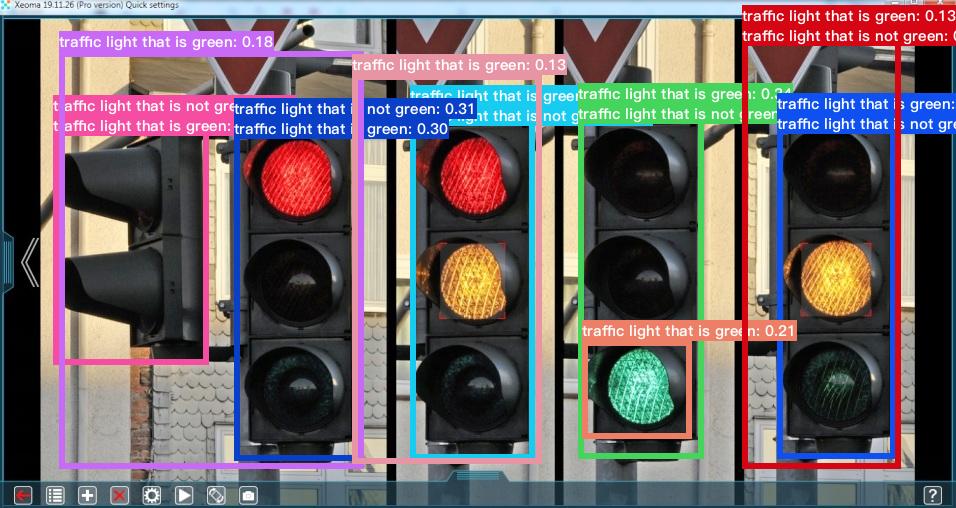}}
    \quad
    \subfigure[]{\includegraphics[width=6.3cm]{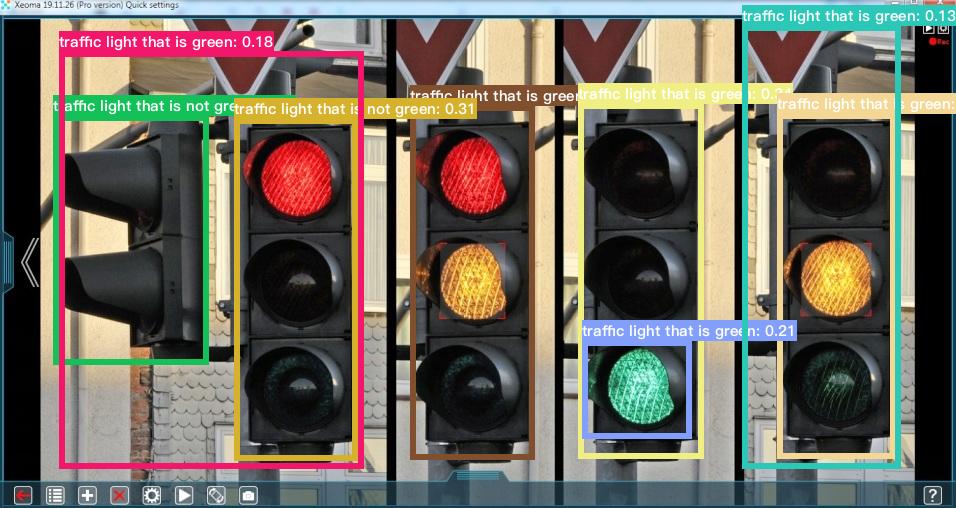}}
    \subfigure[]{\includegraphics[width=6.3cm]{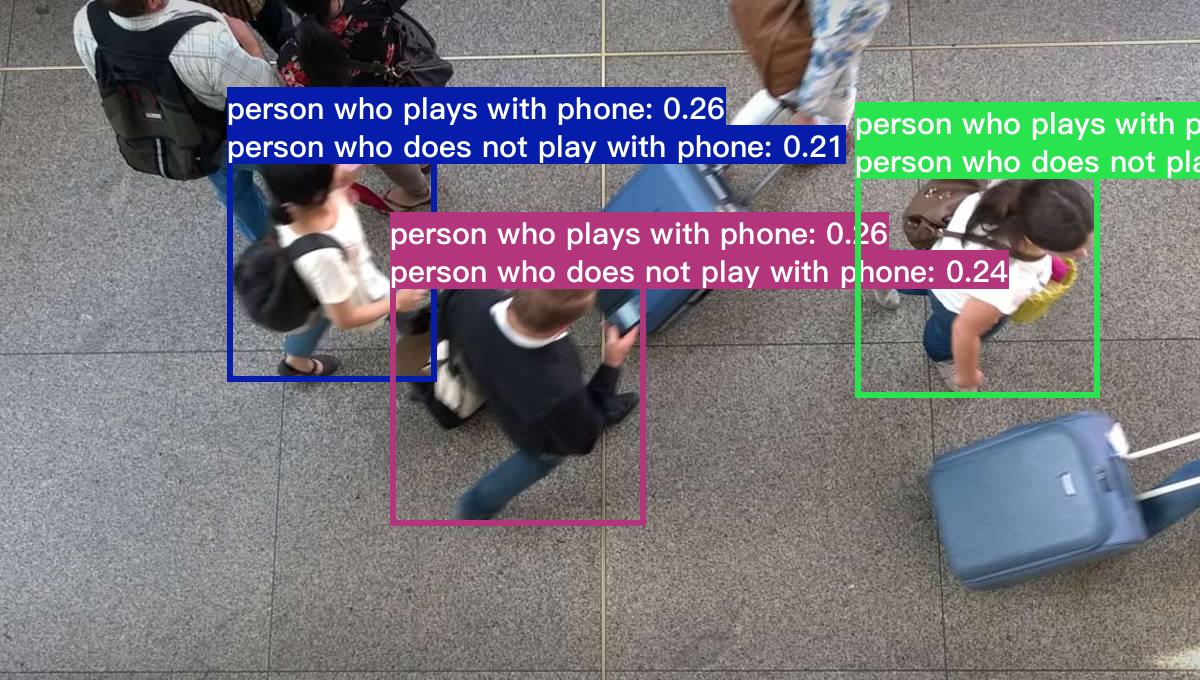}}
    \quad
    \subfigure[]{\includegraphics[width=6.3cm]{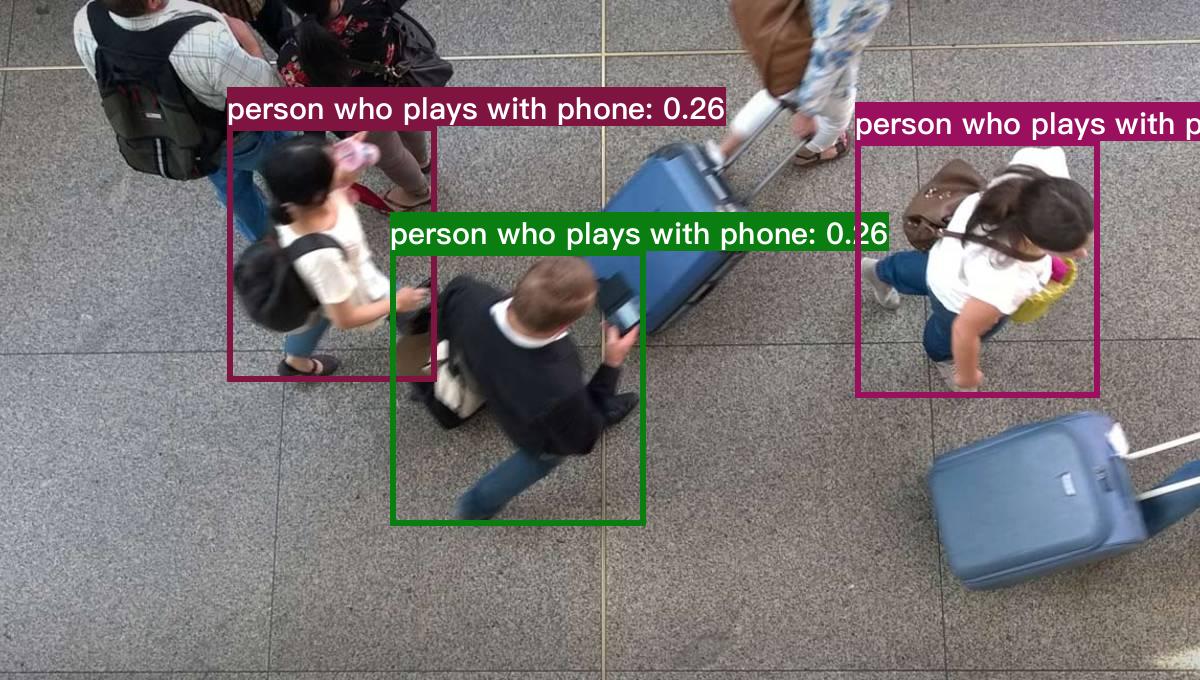}}
\caption{Examples of predictions from GLIP before and after applying class-ignored NMS. The predictions before NMS are presented on the left, while the predictions after NMS are shown on the right.}
\label{fig:examples1}
\end{figure*}

Figure \ref{fig:examples1} presents the prediction results obtained from GLIP using images from the \textit{negation} task. The left column shows visualizations without NMS, while the right column displays visualizations with NMS. In several instances, we observe mutually exclusive labels with bounding boxes on the same object, indicating that OVD models struggle to comprehend detection tasks involving complex descriptions and challenging negative samples. Furthermore, the presence of incorrect labels with higher confidence scores highlights how these models can deceive traditional AP metrics, despite their inability to truly understand the complex descriptions.

\begin{figure}[!ht]
    \centering
    \subfigure[GLIP]{\includegraphics[width=3.1cm]{coco_glip.pdf}}
    \subfigure[GLIP]{\includegraphics[width=3.1cm]{absence_glip.pdf}}
    \subfigure[FIBER]{\includegraphics[width=3.1cm]{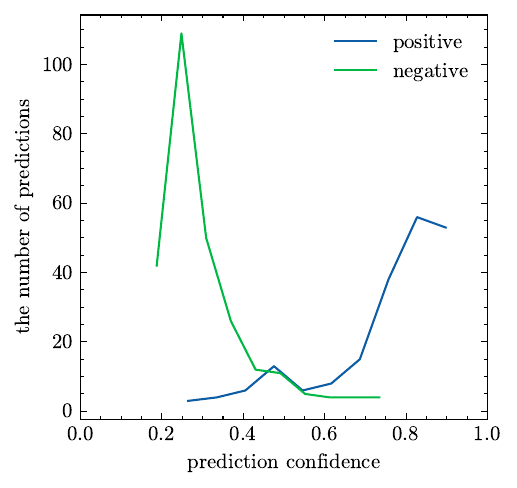}}
    \subfigure[FIBER]{\includegraphics[width=3.1cm]{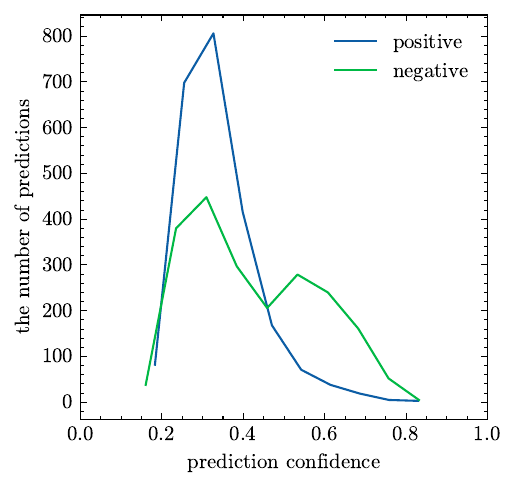}}  
    
    \subfigure[GroundingDINO]{\includegraphics[width=3.1cm]{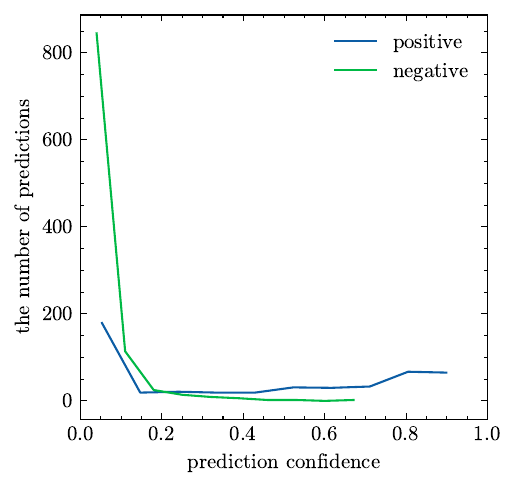}}
    \subfigure[GroundingDINO]{\includegraphics[width=3.1cm]{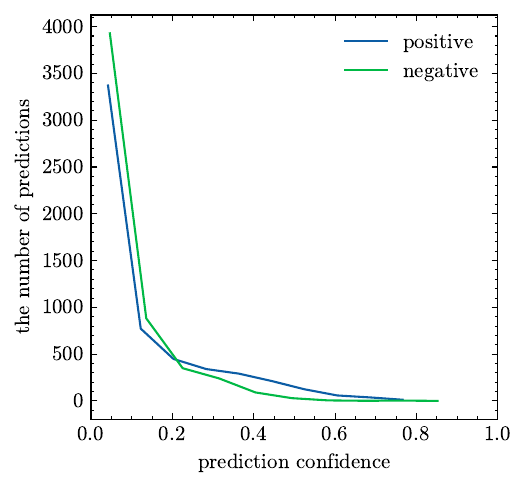}}   
    \subfigure[Detic]{\includegraphics[width=3.1cm]{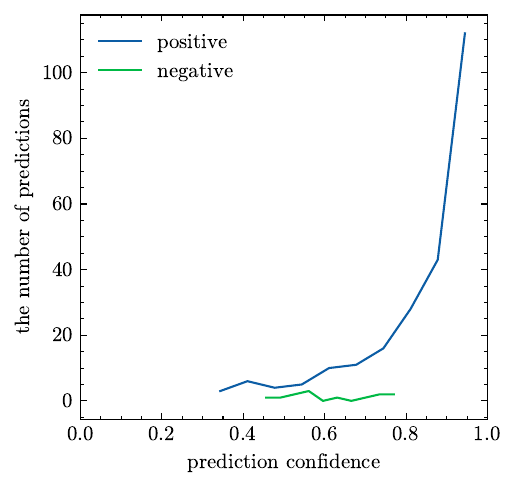}}    
    \subfigure[Detic]{\includegraphics[width=3.1cm]{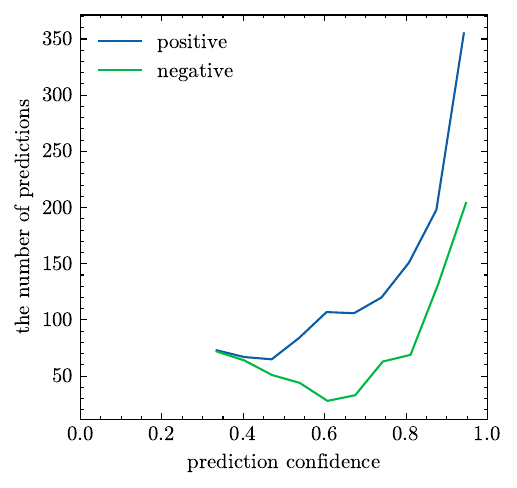}} 
\end{figure}
\begin{figure*}
\centering
    \subfigure[MDETR]{\includegraphics[width=3.1cm]{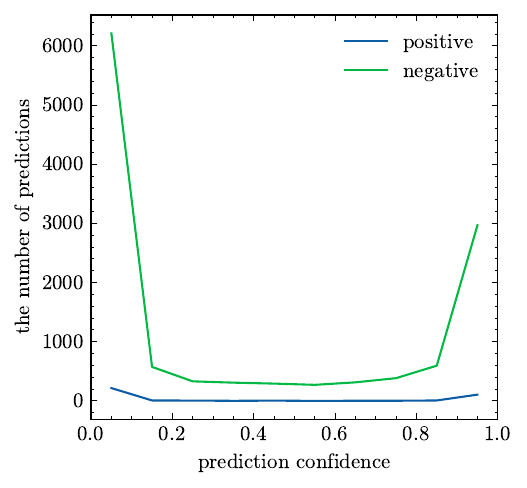}}
    \subfigure[MDETR]{\includegraphics[width=3.1cm]{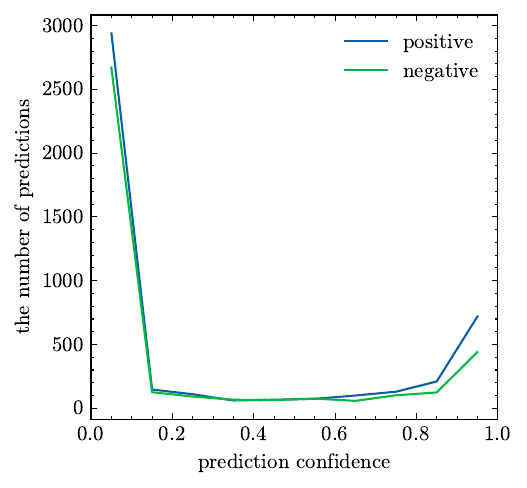}}  
    \subfigure[OmDet]{\includegraphics[width=3.1cm]{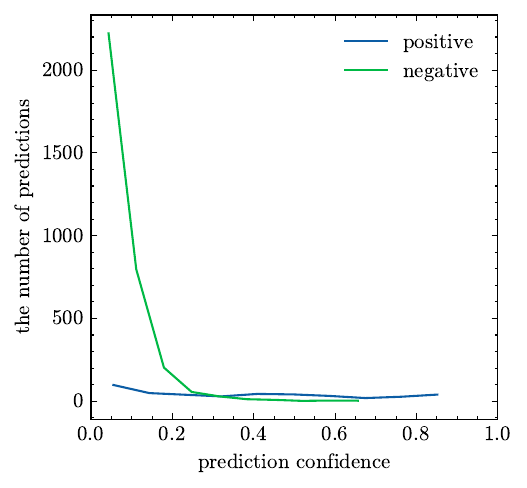}}    
    \subfigure[OmDet]{\includegraphics[width=3.1cm]{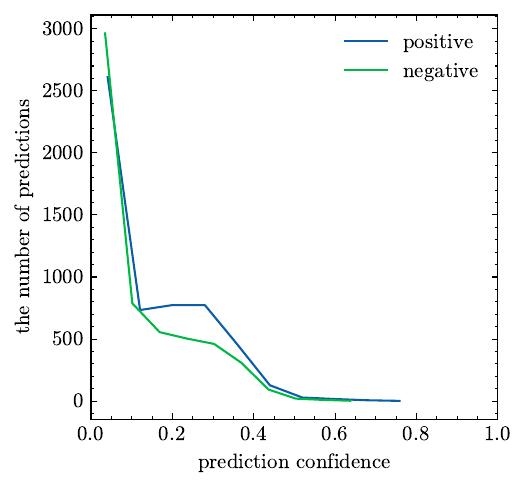}}  
\caption{Distribution analysis of predicted confidence for object and negation aspects in existing OVD Models. (a,c,e,g,i,k) for \textit{object} aspect and (b,d,f,h,j,l) is for \textit{negation} aspect.}
\label{fig:appendix coco-positive and negative}
\end{figure*}

\subsection{Distribution analysis of predicted confidence for object and negation aspects in existing OVD models}

Figure \ref{fig:appendix coco-positive and negative} show the confidence distributions of OVD models. The distributions are drawn for both the \textit{object} and \textit{negation} aspects, enabling a comparison between  common object detection and more complex detection with hard negative samples. This analysis allows us to examine the model predictions and assess their performance in understanding and handling complex descriptions. The distribution of \textit{object} is obtained from the model predictions on a subset of image in COCO validation dataset. Based on the trends observed in the graph for each model, it is evident that they exhibit a strong ability to predict common objects, except for MDETR. Positive predictions are concentrated in the high confidence area, showing high level of certainty. On the other hand, negative predictions are primarily located in the low confidence area, suggesting that the models are confident in classifying these instances as negatives. Indeed, the distribution of the \textit{negation} aspect seems to be mixed up between positive and negative labels, which suggests a failure in the models' understanding and differentiation of positive labels from challenging negative labels with descriptions. This difficulty in accurately classifying and distinguishing between positive and hard negative instances indicates a limitation in the models' ability to comprehend complex descriptions and handle challenging negative samples effectively. 

\end{document}